\pdfoutput=1

\documentclass[11pt]{article}

\usepackage[final]{acl}

\usepackage{times}
\usepackage{latexsym}
\usepackage{amssymb}
\usepackage{soul}   
\usepackage{xcolor}
\usepackage{enumitem}

\usepackage[T1]{fontenc}

\usepackage[utf8]{inputenc}

\usepackage{microtype}

\usepackage{inconsolata}

\usepackage{graphicx}

\usepackage{multirow}
\usepackage{bbding}
\usepackage{amsmath}
\usepackage{tabularx}
\usepackage{booktabs}

\title{\texttt{NUMINA}: A Natural Understanding Benchmark for Multi-dimensional Intelligence and Numerical Reasoning Abilities}

\author{
 \textbf{Changyu Zeng\textsuperscript{1,2,4,}\thanks{Equal contribution.}},
 \textbf{Yifan Wang\textsuperscript{2,}\footnotemark[1]},
 \textbf{Zimu Wang\textsuperscript{1,2,}\footnotemark[1]},
 \textbf{Wei Wang\textsuperscript{1}},
 \textbf{Zhengni Yang\textsuperscript{1,2}}, \\
 \textbf{Muyi Bao\textsuperscript{1,2}},
 \textbf{Jiming Xiao\textsuperscript{1}},
 \textbf{Anh Nguyen\textsuperscript{2}},
 \textbf{Yutao Yue\textsuperscript{3,4,}\thanks{Corresponding author.}}
\\
 \textsuperscript{1}School of Advanced Technology, Xi'an Jiaotong-Liverpool University, China \\
 \textsuperscript{2}Department of Computer Science, University of Liverpool, United Kingdom \\
 \textsuperscript{3}The Hong Kong University of Science and Technology (Guangzhou), China
\\
 \textsuperscript{4}Institute of Deep Perception Technology, JITRI, China
\\
 \texttt{changyu.zeng17@student.xjtlu.edu.cn, yutaoyue@hkust-gz.edu.cn}
}

\begin{document}
\maketitle

\begin{abstract}
Recent advancements in 2D multimodal large language models (MLLMs) have significantly improved performance in vision-language tasks. However, extending these capabilities to 3D environments remains a distinct challenge due to the complexity of spatial reasoning. Nevertheless, existing 3D benchmarks often lack fine-grained numerical reasoning task annotations, limiting MLLMs' ability to perform precise spatial measurements and complex numerical reasoning. To address this gap, we introduce \texttt{NUMINA}, the first \textbf{\underline{N}}atural \textbf{\underline{U}}nderstanding benchmark for \textbf{\underline{M}}ulti-dimensional \textbf{\underline{I}}ntelligence and \textbf{\underline{N}}umerical reasoning \textbf{\underline{A}}bilities to enhance multimodal indoor perceptual understanding. \texttt{NUMINA} features multi-scale annotations and various question-answer pairs, generated using \texttt{NUMINA-Flow}, an automated annotation pipeline that integrates LLM rewriting and rule-based self-verification. We evaluate the performance of various state-of-the-art LLMs on \texttt{NUMINA} following the Chat-Scene framework, demonstrating that current LLMs struggle with multimodal numerical reasoning, particularly in performing precise computations such as distance and volume estimation, highlighting the need for further advancements in 3D models. 
The dataset and source codes can be obtained from \url{https://github.com/fengshun124/NUMINA}.
\end{abstract}


\section{Introduction}
\label{sec:introduction}

Multimodal perception has gained increasing attention in recent years. Leveraging large language models (LLMs), 2D multimodal LLMs (MLLMs) have progressed rapidly, leading to groundbreaking achievements in image classification \cite{zeng2024monotcm,zeng2025skin,bao2025ftcformer} and reasoning \cite{kang2025template,yan2025lmr}.
To extend the applications of MLLMs to 3D domains, researchers have strived to unlock the potential of MLLMs in spatial intelligence to enable practical functionalities, such as autonomous driving \cite{yang2023llm4drive}, indoor navigation \cite{coffrini2025toward}, and route planning \cite{li2025gridroute}. 
Nevertheless, understanding accurate spatial information in 3D environments remains a significant challenge due to the exponential increase in structural complexity introduced. 
For instance, when positioning a cabinet in a living room, one must consider not only the availability of space in the front and back but also factors such as excessive vertical space occupancy and efficient space utilization. Consequently, equipping MLLMs with spatial perception capabilities, including precise measurement and localization, is critical for improving their real-world applicability. 


\begin{figure}
    \centering
    \includegraphics[width=\linewidth]{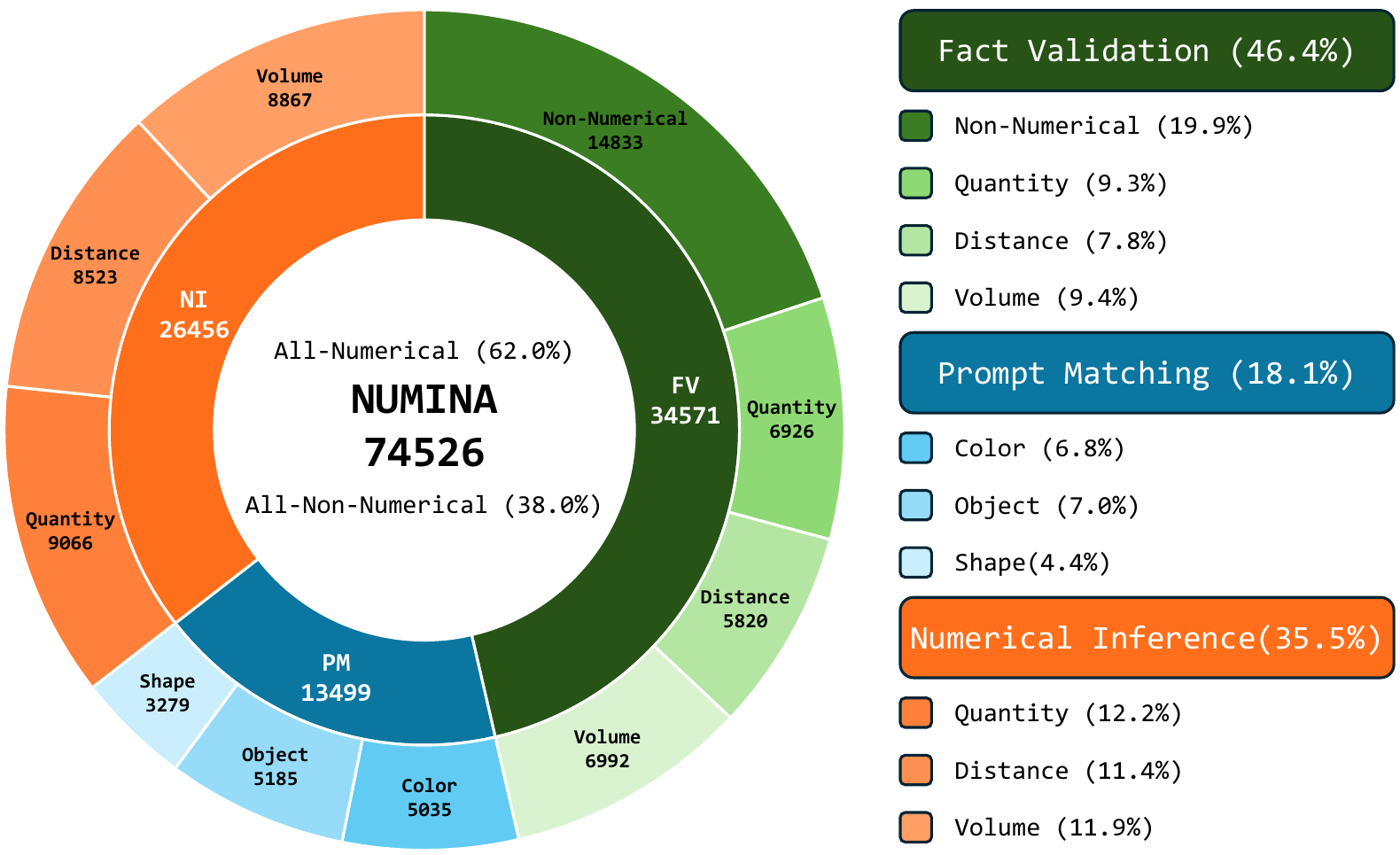}
    \caption{Statistics of the \texttt{NUMINA} benchmark. \texttt{NUMINA} is composed of non-numerical and numerical questions, where the latter are further divided into three categories with increasing difficulty: Fact Validation (FV), Prompt Matching (PM), and Numerical Inference (NI).}
    \label{fig:statistic}
    \vspace{-4mm}
\end{figure}

Achieving precise spatial reasoning necessitates well-annotated 3D vision-language datasets. However, existing benchmarks offer only coarse, global annotations, which limit their applicability for fine-grained reasoning. Datasets like ScanRefer \cite{chen2020scanrefer}, Talk2Nav \cite{vasudevan2021talk2nav}, and ScanQA \cite{azuma2022scanqa} advance 3D tasks by supporting object localization, navigation, and question answering, but they lack multimodal numerical reasoning incorporating spatial precision. Unlike simpler spatial understanding tasks, numerical reasoning involves complex scene interpretation, requiring detailed annotations of object distances, sizes, and positions, which current datasets do not provide. Additionally, these benchmarks predominantly focus on textual outputs, limiting the diversity of question formats and response types, and consequently, their utility in developing models for sophisticated multimodal reasoning. This highlights the need for a more comprehensive, numerically enriched 3D vision-language benchmark.

To this end, we present \texttt{NUMINA}, the first \textbf{\underline{N}}atural \textbf{\underline{U}}nderstanding benchmark for \textbf{\underline{M}}ulti-dimensional \textbf{\underline{I}}ntelligence and \textbf{\underline{N}}umerical reasoning \textbf{\underline{A}}bilities.
As shown in Figure \ref{fig:statistic}, \texttt{NUMINA} stands out by focusing on fine-grained spatial understanding and numerical reasoning in indoor environments, featuring 74,526 question-answer (QA) pairs across three response types with increasing difficulty: Fact Validation (\textit{judgment} questions), Prompt Matching (\textit{multiple-choice} questions), and Numerical Inference (\textit{value-output} questions). Each category is evaluated using tailored metrics and can be further subdivided. These tasks, spanning from simple verification to complex numerical reasoning, ensure that progress on \texttt{NUMINA} reflects genuine advances in 3D multimodal reasoning rather than overfitting to a single question format.

Compared with existing benchmarks, as shown in Table \ref{tab:dataset comparison}, \texttt{NUMINA} is the first dataset that diversely focuses on large-scale, object-level natural understanding of multi-dimensional relationships and numerical reasoning across multiple categories, offering the following key features. (1) \textbf{Comprehensive Numerical Annotations}. Building upon ScanNet \cite{dai2017scannet}, \texttt{NUMINA} introduces multi-dimensional labels, including object center coordinate, bounding box dimensions, and convex hull distances between objects, to enable value-based inference and facilitate the development of diverse numerical reasoning question types. (2) \textbf{Type-Rich Data}. \texttt{NUMINA} enhances global spatial understanding by incorporating LLM rewriting on ScanQA \cite{azuma2022scanqa}, alongside numerical reasoning tasks. The dataset features a diverse task formats, including judgment, multiple-choice, and value-output questions, with difficulty levels ranging from basic to advanced. (3) \textbf{Automatic Annotation Pipeline}. To streamline data generation, we introduce \texttt{NUMINA-Flow}, an automated dataset construction pipeline. \texttt{NUMINA-Flow} employs LLMs to generate high-quality data samples with numerical ground truth.
It also incorporates diverse QA templates, adopts the convex hull distance measurement that aligns with human intuition, and ensures a balanced data distribution to prevent models from exploiting shortcuts or biases.

To rigorously assess the utility of \texttt{NUMINA} and establish a baseline, we utilize several mainstream open-source LLMs, such as Vicuna \cite{chiang2023vicuna} and Qwen \cite{yang2024qwen2}, to serve as decoders within the Chat-Scene \cite{huang2024chat} framework, enabling the simultaneous processing of 3D point clouds, 2D images, and textual inputs. Experimental results reveal substantial challenges of existing LLMs in 3D numerical inference and spatial reasoning, with accuracy falling below 3\% under a 5\% error Threshold Accuracy (TA@5) in distance and volume estimation tasks. While the models perform well in non-numerical evaluations, their accuracy declines considerably in distance-related tasks, achieving only about 54\%, close to random selection. Notably, no single model consistently outperforms across all task categories, highlighting diverse strengths depending on the specific evaluation criteria.

{\renewcommand{\arraystretch}{1.25}
\begin{table}[t!]
    \centering
    \resizebox*{\linewidth}{!}{
    \begin{tabular}{l|c|ccc} \hline
        \multirow{2}{*}{\textbf{Name}} & \multirow{2}{*}{\textbf{\#Pairs}} & \multicolumn{3}{c}{\textbf{Question Types}} \\
        \cline{3-5}
        & & Non-Num. & Num. & Sub-Cat. \\\hline
        ScanRefer & 41,846 & \Checkmark & \XSolidBrush & \XSolidBrush \\
        Scan2Cap & 34,345 & \Checkmark & \XSolidBrush & \XSolidBrush \\
        ScanQA & 30,813 & \Checkmark & \XSolidBrush & \XSolidBrush \\
        {FE-3DGQA} & 20,215 & \Checkmark & \XSolidBrush & \XSolidBrush \\
        SQA3D & 29,884 & \Checkmark & \XSolidBrush & \XSolidBrush \\
        Multi3DRefer & 48,810 & \Checkmark & \XSolidBrush & \XSolidBrush \\
        {CLEVER3D} & 171,174 & \Checkmark & \XSolidBrush & \XSolidBrush \\
        \texttt{NUMINA} & 74,526 & \Checkmark & \Checkmark & \Checkmark \\\hline
        
    \end{tabular}}
    \caption{Comparison between \texttt{NUMINA} and existing 3D vision-language benchmarks. (Num.: Numerical Questions; Cat.: Categories)}
    \label{tab:dataset comparison}
    \vspace{-4mm}
\end{table}
}
\section{Related Work}
\label{sec:related work}

\begin{figure*}[t!]
    \centering
    \includegraphics[width=0.95\linewidth]{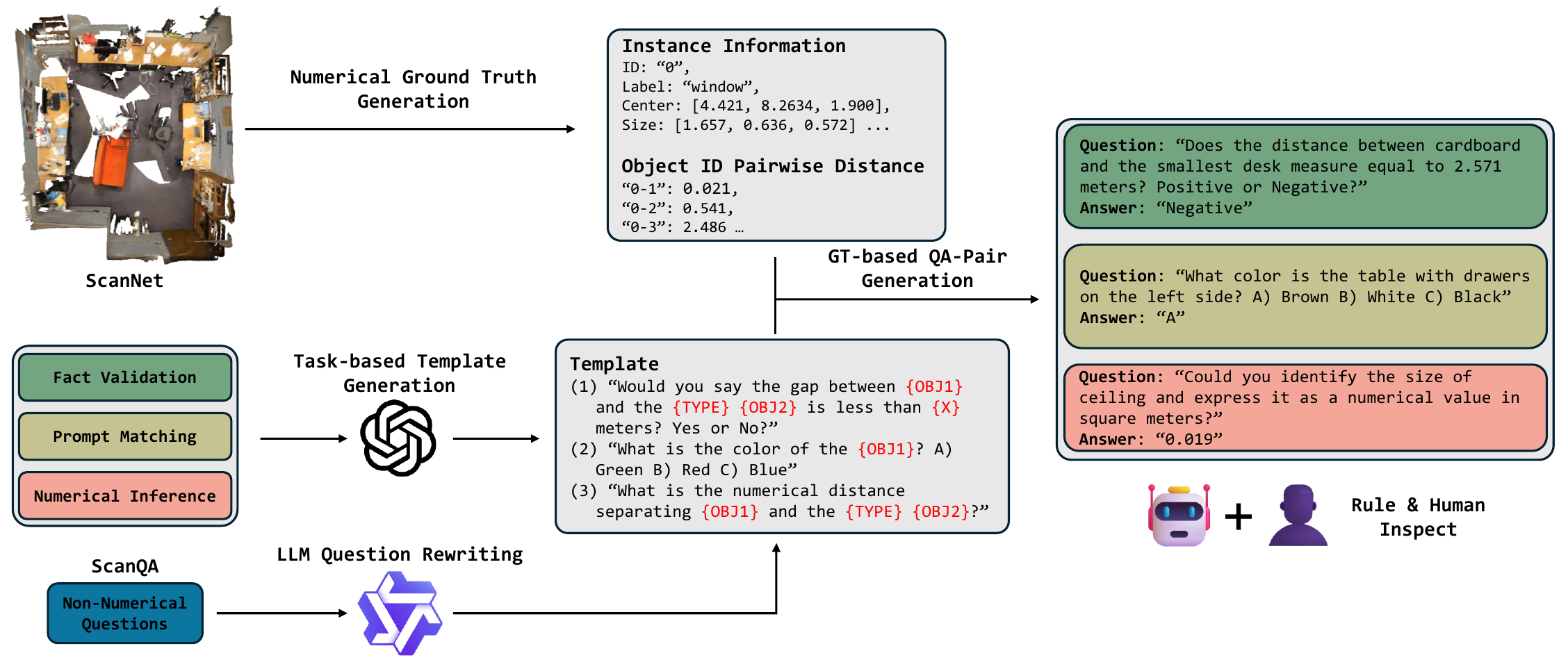}
    \caption{Overview of the \texttt{NUMINA-Flow} pipeline. Numerical Ground Truth (NGT) is extracted from ScanNet, including instance details and pairwise distances. GPT-4o generates diverse question templates filled with NGT, followed by rule-based and manual validation. Non-numerical questions are rewritten using Qwen2.5-72B with the ScanQA dataset for added diversity.
    }
    \label{fig:pipeline}
    \vspace{-4mm}
\end{figure*}

\paragraph{3D Vision-Language Understanding.}

Existing 3D vision-language benchmarks play a pivotal role in advancing research on vision and natural language understanding in 3D environments. ScanRefer \cite{chen2020scanrefer} is a pioneering effort, integrating detailed textual descriptions with complex indoor scenes to facilitate 3D object localization based on natural language queries. Building on this foundation, ScanQA \cite{azuma2022scanqa} expands the scope by introducing diverse QA pairs to support various vision-language tasks, such as 3D visual grounding and 3D visual question answering (VQA). ReferIt3D \cite{achlioptas2020referit3d} focuses on fine-grained 3D object localization using natural language references, encompassing both real-world (Natural ReferIt3D, Nr3D) and synthetic (Synthetic ReferIt3D, Sr3D) contexts. Meanwhile, Talk2Nav \cite{vasudevan2021talk2nav} introduces dialog-based interactions with agents, enabling the clarification of ambiguous or incomplete instructions through natural language queries, particularly for long-range navigation tasks. However, none of these benchmarks offer multi-level numerical annotations, which poses challenges for training models capable of achieving precise 3D perception. As a result, such models face limitations in applications requiring accurate measurement and localization, such as interior design and spatial planning.


\paragraph{Indoor Environment Perception.}
Indoor environment perception is an essential research field with broad applications in robotics and augmented reality (AR). Researchers have proposed various approaches leveraging diverse sensing modalities and computational methods to advance scene understanding and mapping. Vision-based techniques \cite{ran2021scene,ruotsalainen2021improving}, which rely on RGB or RGB-D cameras, often employ deep neural networks (DNNs) such as R-CNN \cite{he2017mask} and U-Net \cite{ronneberger2015u,cao2022swin} for tasks like object detection and semantic segmentation, enabling the identification of structural components within indoor spaces \cite{gan2025review}. However, the limitations of these systems, such as visual illusions and occlusions caused by the absence of depth information, have led to the adoption of LiDAR and depth cameras. These sensors provide rich 3D spatial information, enabling methods like Cartographer \cite{dwijotomo2020cartographer} and KinectFusion \cite{izadi2011kinectfusion} to construct accurate geometrical representations of indoor environments. Despite notable advancements, several challenges remain. For instance, current MLLMs often lack the precise numerical reasoning capabilities required to locate objects accurately within a room. Additionally, there is a significant shortage of indoor multimodal datasets with comprehensive, multi-level numerical annotations, hindering the development of robust perception models.

\section{\texttt{NUMINA} and \texttt{NUMINA-Flow}}
\label{sec:benchmark}


\begin{figure*}[t!]
    \centering
    \includegraphics[width=0.95\linewidth]{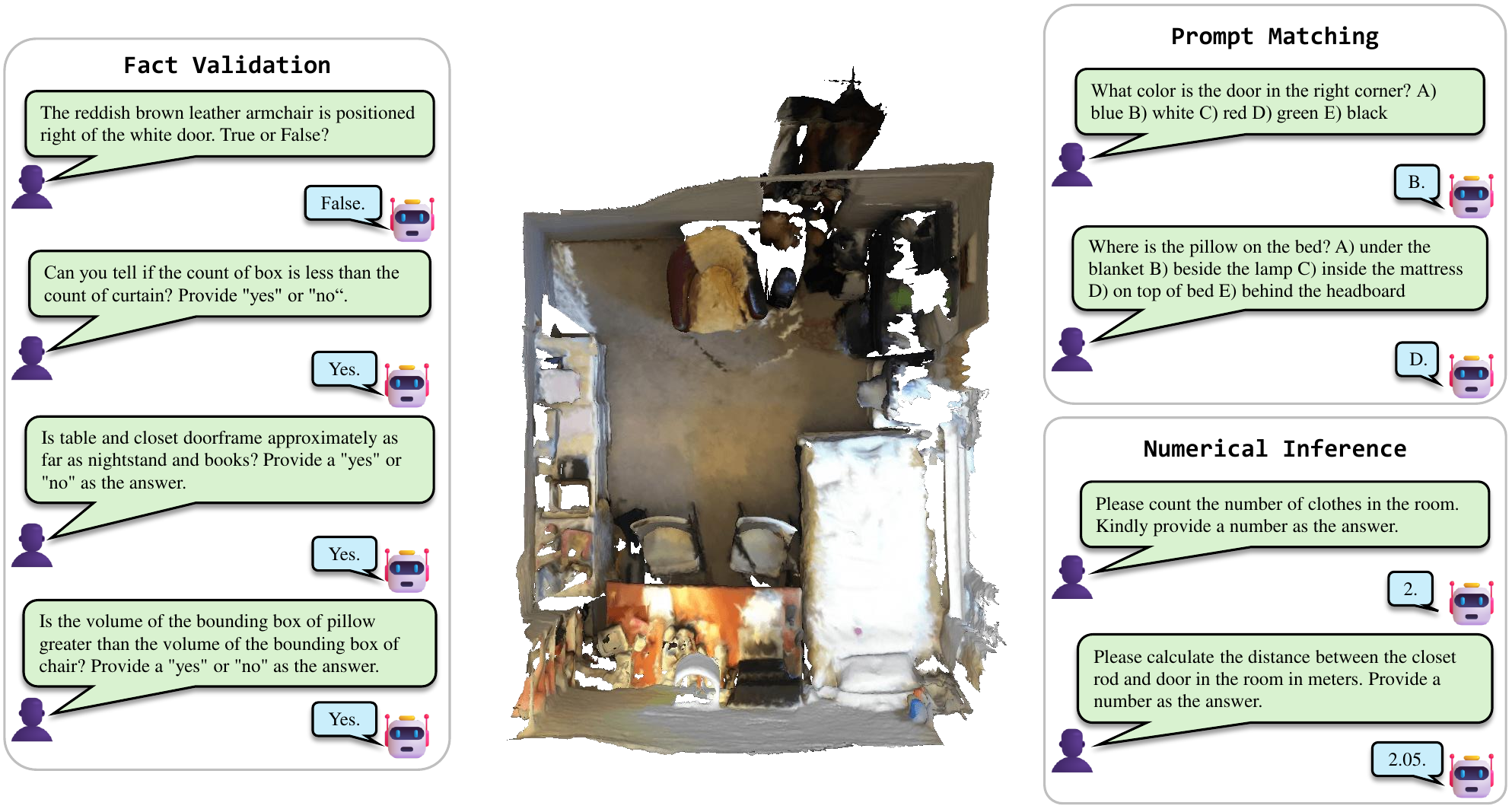}
    \caption{Example of various scene understanding and numerical reasoning tasks in \texttt{NUMINA} dataset. All tasks are formulated as single-turn question-answering pairs without the use of additional task-specific heads, ensuring a unified and consistent evaluation framework.
    }
    \label{fig:demo}
    \vspace{-2mm}
\end{figure*}

This section describes in detail the automatic construction pipeline \texttt{NUMINA-Flow} for the \texttt{NUMINA} dataset. As illustrated in Figure \ref{fig:pipeline}, \texttt{NUMINA-Flow} generates numerical data by extracting Numerical Ground Truth (NGT) from ScanNet \cite{dai2017scannet}. It leverages GPT-4o \cite{hurst2024gpt}  to create question-answer templates, {followed by rule-based and manual verification. The generated templates are validated by human evaluators to ensure clarity and ambiguity, making them comprehensible to both humans and models. Additionally, variations in expression and syntactic structure are incorporated to promote generalization across different reasoning patterns.} To enhance question diversity, non-numerical questions are further generated using Qwen2.5-72B \cite{yang2024qwen2}.

\subsection{Dataset Construction} \label{subsec:Dataset Construction}
To bootstrap the dataset construction process, we select the indoor 3D scenes from the existing ScanNet \cite{dai2017scannet} dataset. Building on these complex indoor environments, \texttt{NUMINA} expands the scope of questions beyond text-only answers. Three distinct tasks, designed with varying levels of difficulty, are defined as follows. Examples of these tasks are shown in Figure \ref{fig:demo}.





\begin{itemize}[itemindent=0em,itemsep=0em]
    \item \textbf{Fact Validation (FV):} Binary classification questions requiring ``yes'' or ``no'' responses to evaluate factual accuracy.
    \item \textbf{Prompt Matching (PM):} Multiple-choice questions requiring the selection of the correct answer from five options, demanding stronger spatial comprehension.
    \item \textbf{Numerical Inference (NI):} The most challenging task requiring precise numerical outputs for quantities, volumes, or distances.
\end{itemize}

To enhance diversity beyond numerical reasoning, 14,000 QA pairs from ScanQA \cite{azuma2022scanqa} covering color, shape, and spatial relationships are integrated after rewriting with Qwen2.5-72B \cite{yang2024qwen2}, with diverse templates in Appendix \ref{sec:templates}. This yields the dataset of 74,526 total QA pairs, containing 46,194 numerical QA pairs (62.0\%) across quantity, distance, and volume subcategories, plus 28,332 non-numerical pairs (38.0\%) from ScanQA, as shown in Figure \ref{fig:statistic}.

\subsection{Numerical Ground Truth Extraction} \label{subsec:NGT_generation}
Based on the original annotations, we calculate crucial geometric parameters for each instance in the scene, including centroid coordinates, bounding box dimensions, and the minimum and maximum values along the $(x,y,z)$ axes. Furthermore, we calculate the convex hull distance \cite{bentley1982approximation} for every pair of instances. This metric, defined as the shortest distance between the outer boundaries of two point sets determined by their convex hulls, provides a precise representation of relative positional relationships, especially for complex shapes, as it accounts for both the center points and the overall geometry. We adopt this distance due to its strong alignment with human visual perception \cite{bentley1982approximation}. To represent the volume of irregular point cloud objects, we utilize the smallest axis-aligned bounding box, a rectangular box aligned with the coordinate axes that fully encloses the object. The calculated pairwise convex hull distances and bounding box dimensions are systematically organized and form the basis for generating QA pairs in subsequent stages.

\subsection{QA-Pair Generation} \label{subsec:QA_generation}

Our QA-pair generation process follows a systematic three-stage pipeline designed to ensure linguistic diversity, factual accuracy, and balanced difficulty distribution.

\paragraph{Step 1: Template Generation and Selection.} To enhance linguistic diversity and question complexity, we employ GPT-4o \cite{hurst2024gpt} to generate ten distinct question templates for each task category. These templates incorporate varied syntactic structures and expressions to promote generalization across different reasoning patterns. During question generation, templates are randomly sampled to ensure diversity, with reserved placeholders systematically replaced by NGT values and object references from the target scene.

\paragraph{Step 2: Task-specific Formatting.} Each task type requires specific formatting to ensure clear evaluation criteria. For Fact Validation (FV) tasks, we append explicit prompts such as ``\textit{Is the statement Yes or No?}'' to ensure binary output. Prompt Matching (PM) tasks include constraint-oriented instructions guiding models to select from five given options (from A to E). Numerical Inference (NI) tasks explicitly request precise numerical values with appropriate units.

\paragraph{Step 3: Bias Mitigation and Balance.} To prevent systematic biases, we implement several balancing strategies: (1) For PM tasks, correct answers are uniformly distributed across all five options (A-E) to prevent positional bias; (2) For FV tasks, ``yes'' and ``no'' answers are equally represented; (3) Template selection follows uniform random sampling to ensure no linguistic patterns dominate the dataset.

Furthermore, to expand dataset diversity beyond purely numerical reasoning, we integrate 14,000 non-numerical QA pairs from ScanQA \cite{azuma2022scanqa} covering color, shape, and spatial relationships. These questions undergo systematic rewriting using Qwen2.5-72B \cite{yang2024qwen2} to conform to our FV and PM task formats, ensuring consistency with our evaluation framework while preserving their original semantic content.

The complete pipeline produces 74,526 QA pairs with progressive difficulty levels across all categories. Detailed templates and examples for each task type are provided in Appendix \ref{sec:templates}.

\subsection{Rule-based and Human Inspection} \label{subsec:inspection}

To ensure \texttt{NUMINA}’s accuracy and consistency, outputs from the automatic \texttt{NUMINA-Flow} pipeline undergo a two-stage validation. First, a rule-based framework detects issues in Qwen2.5-generated questions, such as hallucinations or incorrect option counts, and regenerates faulty items up to five times. Questions must meet all inspection criteria before passing this stage. Second, five human evaluators conduct a detailed review of 20,000 randomly selected samples, checking for grammar, logic, expression, and unit consistency, achieving a correctness rate of 99.5\%, validating the quality and reliability of the \texttt{NUMINA} dataset. 



All aforementioned annotation processes, except human inspection, are consolidated and integrated into the single \texttt{NUMINA-Flow} annotation pipeline. This pipeline enables practitioners to replicate and further improve our annotation procedure or modify specific portions of the code to create customized datasets tailored to their needs. The detailed process of the QA-pair generation is presented in Appendix \ref{sec:details}.


\begin{figure*}[t!]
    \centering
    \includegraphics[width=\linewidth]{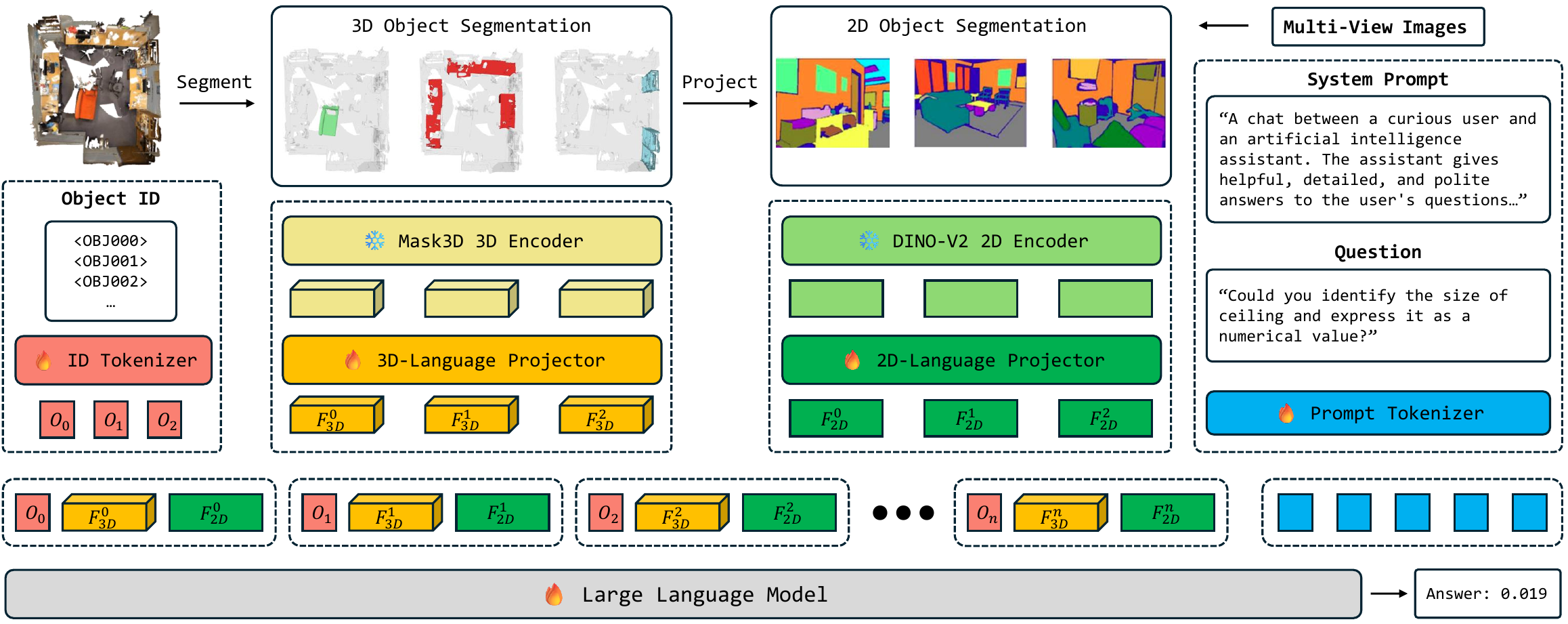}
    \caption{Overall Chat-Scene framework. The framework processes 3D scenes through a multi-stage pipeline: (1) scene decomposition into object segments, (2) mapping of segments to multi-view images via corresponding masks, (3) extraction of object-centric representations using dedicated 3D and 2D encoders, and (4) combination of these representations with object identifiers to generate scene embeddings as sequences of object-level embeddings. These embeddings serve as input to the language model component. For evaluation on the NUMINA benchmark, we substitute the original large language model with open-source alternatives including Vicuna, Mistral, Qwen, and Phi to assess their numerical reasoning capabilities.
    }
    \label{fig:model}
\end{figure*}





\section{Experiments}\label{sec:experiments}

\subsection{Model Architecture}

To assess data quality and task difficulty, we conduct our 3D-language VQA experiments following the Chat-Scene \cite{huang2024chat} paradigm. As illustrated in Figure \ref{fig:model}, a pre-trained detector segments the point cloud into individual objects, which are then projected onto multi-view images using corresponding masks. Object-centric features are extracted using tailored 3D Mask3D \cite{schult2023mask3d} and 2D DINO-V2 \cite{oquab2023dinov2} encoders, processed through projection layers, and integrated with unique object identifiers (IDs) to form a sequence of object-level embeddings. Finally, given the object-level embeddings and question prompts, the LLM generates a concise response either True or False, a multiple-choice option, or a numerical value based on the task type.  

\subsection{Evaluation Metrics}
Since the outputs in all three tasks of \texttt{NUMINA} are straightforward and unambiguous, we define distinct metrics for each task separately. For the Fact Validation (FV) and Prompt Matching (PM) tasks, we use accuracy as the evaluation metric, where a response is considered correct only if the prediction exactly matches the ground truth. For the Numerical Inference (NI) task, accurately measuring the distance between two objects or the size of a specific item presents a significant challenge for the model. Therefore, we employ Threshold Accuracy (TA) as the assessment metric, which measures the model's performance within specific pre-defined thresholds. Formally, the TA metric can be expressed as follows:
\begin{equation}
    \text{TA} = \frac{1}{N} \sum_{i=1}^N \mathbf{1}\left(|d_i^{\text{pred}} - d_i^{\text{true}}| < \text{Threshold}\right),
\end{equation}
where $d_i^{\text{pred}}$ represents the model's prediction, and $d_i^{\text{true}}$ denotes the ground truth. The indicator function $\mathbf{1}$ evaluates to 1 if the specified condition is satisfied and 0 otherwise. Threshold is a predefined value that limits the allowable error between the prediction and the ground truth. In our case, we set the threshold values as 5\%, 10\%, 20\% corresponding to TA@5, TA@10, TA@20, respectively.

{\renewcommand{\arraystretch}{1.25}
\begin{table*}[htpb]
    \centering
    \resizebox*{\linewidth}{!}{
    \begin{tabular}{l|c|cccc|ccc} \hline
        \multirow{2}{*}{\textbf{LLM}} & \multicolumn{1}{c|}{\textbf{Prompt Matching}} & \multicolumn{4}{c|}{\textbf{Fact Validation}} & \multicolumn{3}{c}{\textbf{Numerical Inference}} \\
        \cline{2-9}
        & Non-Num & Non-Num & Quantity & Distance & Volume & Quantity & Distance & Volume  \\\hline
        Vicuna 7B v1.5 & 83.35 & 77.31 & 71.69 & 53.89 & 83.44 & 8.01 & 1.73 & 2.40 \\
        Vicuna 13B v1.5 & 85.12 & 76.17 & 71.30 & 54.35 & 83.66 & 10.15 & 2.71 & 3.06 \\
        {Phi 4-mini} & 84.79 & 75.24 & 65.23 & 51.62 & 80.36 & 10.36 & 1.21 & 3.17 \\
        {Mistral 7B} & 84.75 & 75.88 & 71.97 & 55.31 & 83.14 & 8.49 & 2.54 & 3.22 \\
        Qwen2.5-7B & 85.70 & 75.34 & 69.76 & 54.29 & 82.95 & 10.63 & 1.21 & 1.80 \\
        DeepSeek-R1-Distill-Qwen-7B & 82.19 & 73.11 & 69.59 & 52.70 & 81.81 & 9.99 & 2.02 & 2.67 \\\hline
    \end{tabular}}
    \caption{Performance comparison of different LLMs on the \texttt{NUMINA} benchmark for different kinds of problems. PM and FV are evaluated by Accuracy. NI is assessed by the Threshold Accuracy (TA@5), which measures the proportion of predictions falling within a 5\% error margin with respect to the ground truths.}
    \vspace{-4mm}
    \label{tab:performance comparison}
\end{table*}
}


\subsection{Experimental Setup}
{We adhere to the original ScanNet} \cite{dai2017scannet} {split, using the training set for training and the validation set for evaluation, as the test set annotations are not publicly available. This protocol is consistent with prior works such as ScanQA} \cite{azuma2022scanqa}. {All tasks are formulated within a user–assistant interaction framework. Joint training is performed using the Cross-Entropy loss function and the AdamW optimizer, enabling parameter updates across unfrozen components of the model, including the tokenizer, projector, and LLMs.}

We adopt six open-source LLMs in our experiments: Vicuna 7B v1.5, Vicuna 13B v1.5 \cite{chiang2023vicuna}, Qwen2.5-7B \cite{yang2024qwen2}, {Phi 4-mini} \cite{abouelenin2025phi}, {Mistral-7B} (\texttt{Instruct-v0.3}, \citealp{jiang2023mistral7b}) and DeepSeek-R1-Dstill-Qwen-7B \cite{guo2025deepseek}. To fine-tune these models, we apply LoRA \cite{hu2021lora} with a rank of 16. The learning rate is set to 2e-6, following a cosine annealing schedule. The training process spans three epochs and is completed using 4 NVIDIA A100 Tensor Core GPUs.

\subsection{Experiment Results}
Table \ref{tab:performance comparison} presents the performance of the six LLMs across various \texttt{NUMINA} subtasks. In Prompt Matching (PM), all models achieve high accuracy on non-numerical tasks, ranging from 82.19\% to 85.70\%. Similarly, in Fact Validation (FV), models perform well on non-numerical assessments but exhibit a significant drop in accuracy for distance-related tasks, with scores around 54\%, indicating challenges in understanding convex hull distances. Numerical Inference (NI) proves to be the most difficult category, particularly in distance and volume estimation, where accuracy remains below 3\%. Among the models, Vicuna 13B v1.5 achieves the highest average score in numerical inference, though it only slightly outperforms Vicuna 7B v1.5 by 1.26\%. These results highlight the limitations of current LLMs in fine-grained 3D spatial reasoning, particularly in tasks requiring precise numerical analysis. For Qwen, Phi, Mistral, and the latest DeepSeek-R1 distilled models, their performance has been unsatisfactory. Overall, no single model excels in all numerical sub-tasks. These six large models exhibit varying strengths and weaknesses depending on the specific tasks and problems to which they are applied. Figure \ref{fig:demo} provides some illustrative examples of the three tasks for further clarification.

\subsection{Analysis}

Current MLLMs exhibit strong performance on purely linguistic and non‐numerical spatial tasks (Prompt Matching and non‐numeric Fact Validation, both >75\% accuracy) but break down when asked to perform fine‐grained numerical reasoning, especially in 3D settings. Architecturally, Transformers lack inductive biases for geometry: they treat numbers as discrete tokens without true magnitude awareness and are not designed to encode spatial relationships such as distances, angles, or volumes. Moreover, standard pretraining corpora of LLMs seldom include explicit 3D‐spatial supervision or grounded numerical examples, leaving models underexposed to the types of geometric computations required by \texttt{NUMINA}.

Even in emerging multimodal variants, MLLMs typically fuse visual inputs only via generic embeddings, without dedicated modules for point‐cloud processing or multi‐view depth estimation. As a result, tasks like convex‐hull distance estimation and precise volume calculation, which demand multi‐step geometric reasoning and spatial continuity, fall outside the models’ learned capabilities. The anomalously high “volume” accuracy (83\%) likely reflects reliance on commonsense size priors (e.g., “\textit{monitors are larger than phones}”) rather than true geometric inference.

These findings suggest that improving numerical reasoning in MLLMs requires fundamental architectural changes beyond simply scaling model size. The consistent poor performance across all tested models (including recent architectures like Phi4 \cite{abouelenin2025phi} and DeepSeek-R1 \cite{guo2025deepseek} indicates these limitations reflect deeper constraints in how current transformer architectures process spatial and numerical information, pointing toward the need for specialized geometric reasoning modules and grounded training paradigms that incorporate explicit 3D spatial supervision.

\subsection{Ablation Study} \label{subsec:ablation}
\paragraph{Logical Consistency.}

{\renewcommand{\arraystretch}{1.25}
\begin{table}[t]
    \centering
    \resizebox*{0.9\linewidth}{!}{
    \begin{tabular}{c|c|ccc} \hline
        \multirow{2}{*}{\textbf{LLM}} & \multirow{2}{*}{\textbf{Type}} & \multicolumn{3}{c}{\textbf{Fact Validation}} \\ \cline{3-5}
        & & Quantity & Distance & Volume  \\\hline
        \multirow{2}{*}{Vicuna 7B v1.5} & Ori &  74.47 & 65.52 & 63.64 \\ & CP  & 73.56 & 67.06 & 66.32 \\\hline
        \multirow{2}{*}{Vicuna 13B v1.5} & Ori & 73.61 & 64.69 & 64.88 \\ & CP & 74.47 & 66.23 & 63.84 \\\hline
        \multirow{2}{*}{Qwen2.5-7B} & Ori & 72.49 & 67.18 & 72.11 \\ & CP & 74.89 & 69.08 & 72.31 \\\hline
    \end{tabular}}
    \caption{Logical consistency ablation study. ``Ori'' refer to original QA-pair and ``CP'' indicates the corresponding contrapositive ones.}
    \label{tab:lc}
    \vspace{-4mm}
\end{table}
}

The Fact Validation (FV) task includes two sets of questions with corresponding answers designed to evaluate semantic understanding, common sense reasoning, and logical coherence. For example, in one QA pair, the original question asks, ``\textit{Is it correct that the sofa is bigger than the cell phone?}'', with the expected answer being ``\textit{Yes.}'' In the corresponding contrapositive pair, the question states, ``\textit{The sofa is smaller or equal than the cell phone}'', with the correct answer being ``\textit{No.}'' By systematically altering relational statements and their corresponding answers, we test the model’s ability to maintain logical consistency and accurately interpret semantic relationships. Details are in Appendix \ref{sec:lc}.

Table \ref{tab:lc} summarizes the logical consistency evaluation results of three models in the FV task. The quantitative analysis reveals that the average performance discrepancy between original and contrapositive QA-pairs across these models is 1.71\%, 1.15\%, and 1.50\%, respectively. Notably, Vicuna 13B v1.5 exhibits superior performance stability compared to other models, which can be reasonably attributed to its enhanced model capacity with 13 billion parameters. Experimental findings demonstrate that all evaluated LLMs exhibit robust logical reasoning capabilities, which consistently produce accurate judgments, even when the logical structure of both the questions and corresponding answers is reversed. This outcome highlights the models’ resilience to logical perturbations and their capacity to generalize reasoning patterns beyond fixed, predefined formats. Furthermore, model scale appears to provide a stronger baseline for performance. However, the training methodology and the nature of the source data, such as differences between Vicuna and Qwen, play a critical role in enabling more complex forms of reasoning. These factors suggest that both architectural scale and training data composition significantly influence the depth and flexibility of logical reasoning in LLMs.

\paragraph{Effects of Chain-of-Thought.}
\begin{table}[t]
    \centering
    \resizebox*{0.9\linewidth}{!}{
    \begin{tabular}{c|ccc} \hline
        \multirow{2}{*}{\textbf{LLM}} & \multicolumn{3}{c}{\textbf{Fact Validation}} \\
        & Quantity & Distance & Volume  \\\hline
        Vicuna 7B v1.5 & 63.41 & 53.76 & 80.72 \\
        \multicolumn{1}{r|}{+ CoT} & 66.61 & 56.93 & 79.85 \\\hline
        Vicuna 13B v1.5 & 70.14 & 54.55 & 83.12 \\
        \multicolumn{1}{r|}{+ CoT} & 72.20 & 55.34 & 79.74 \\\hline
    \end{tabular}}
    \caption{Chain-of-Thought study on fact validation.}
    \label{tab:cot-fv}
\end{table}

\begin{table}[t]
    \centering
    \resizebox*{0.9\linewidth}{!}{
    \begin{tabular}{c|ccc} \hline
        \multirow{2}{*}{\textbf{LLM}} & \multicolumn{3}{c}{\textbf{Numerical Inference (Volume)}} \\
         & TA@5 & TA@10 & TA@20  \\\hline
        Vicuna 7B v1.5 & 2.62 & 4.91 & 8.41 \\
        \multicolumn{1}{r|}{+ CoT} & 2.95 & 4.20 & 11.95 \\\hline
        Vicuna 13B v1.5 & 2.67 & 4.48 & 8.57 \\
        \multicolumn{1}{r|}{+ CoT} & 3.88 & 5.46 & 9.39 \\\hline
    \end{tabular}}
    \caption{Chain-of-Thought study on volume numerical inference task. TA@5/10/20 indicates Threshold Accuracy at error levels of 5\%, 10\%, and 20\%.}
    \label{tab:cot-ni}
    \vspace{-4mm}
\end{table}

In many natural language tasks, the Chain-of-Thought (CoT) paradigm has been shown to enhance reasoning by guiding models through step-by-step problem-solving processes. To investigate its effectiveness in value reasoning tasks, we adapted FV and volume-related tasks of the NI problems to the CoT framework. Specifically, instead of providing short numerical answers, we incorporate structured intermediate reasoning steps leading to the final answers. For instance, in volume-related questions, the CoT paradigm first extracts the 3D bounding box data of the specified object. It then computes the object's volume by multiplying the bounding box dimensions, ultimately deriving the final size. This structured reasoning process aims to improve the model’s numerical inference ability. A concrete example can be found in Appendix \ref{sec:cot}.

Tables \ref{tab:cot-fv} and \ref{tab:cot-ni} present the impact of CoT reasoning on Vicuna 1.5 models across FV and NI tasks. The results reveal mixed effectiveness of CoT across different model scales and task types. For the 7B model, CoT demonstrates consistent improvements in quantity (+3.20\%) and distance (+3.17\%) estimation within FV tasks, while showing minimal degradation in volume estimation (-0.87\%). Conversely, the 13B model exhibits more variable performance: improvements in quantity (+2.06\%) and distance (+0.79\%) tasks are accompanied by a notable decline in volume estimation (-3.38\%).
In the NI task, both models show modest improvements with CoT. The 7B model achieves gains across all threshold levels, with the most substantial improvement at TA@20 (+3.54\%). Similarly, the 13B model demonstrates improvements at TA@5 (+1.21\%) and TA@10 (+0.98\%), though with a slight decrease at TA@20 (-1.18\%).

The varied effectiveness of CoT can be attributed to a complex interplay between task requirements and the reasoning process it induces. CoT demonstrates significant utility for tasks amenable to logical decomposition, such as quantity and distance estimation, by enforcing a structured, procedural pathway. In contrast, its efficacy diminishes for tasks that rely on holistic, commonsense heuristics, as exemplified by the performance degradation in volume fact validation. This negative impact likely stems from CoT's explicit reasoning steps interfering with the models' more accurate, pre-existing intuitive knowledge, an effect amplified in the larger 13B model, whose robust implicit knowledge is more susceptible to such disruption. 



\section{Conclusion}

\label{sec:conclusion}
We introduce \texttt{NUMINA}, a Natural Understanding Benchmark for Multi-dimensional Intelligence and Numerical Reasoning Abilities, which is designed to advance multimodal indoor perceptual understanding. \texttt{NUMINA} provides a diverse dataset with multi-scale annotations and a variety of question-answer pairs spanning multiple task types. This dataset is generated using \texttt{NUMINA-Flow}, an automated annotation platform that leverages LLM rewriting and rule-based self-verification. Following the Chat-Scene paradigm, we replace LLMs within the architecture and jointly train and evaluate all tasks. Experimental results indicate that current LLMs exhibit significant limitations in multimodal numerical reasoning, particularly in tasks requiring fine-grained computational precision, such as quantity, distance, and volume estimation. 
In the future, we will develop further advancements in numerical and spatial reasoning for multimodal AI systems.


\section*{Limitations}
While the \texttt{NUMINA} benchmark represents a significant advancement in the assessment of multi-dimensional intelligence and numerical reasoning, several limitations warrant consideration. First, the complexity of the benchmark necessitates substantial computational resources, which may pose accessibility challenges for researchers with limited hardware capabilities. Second, although the benchmark encompasses a wide range of tasks, it may not fully capture all aspects of human cognition, particularly those dimensions that are inherently subjective or difficult to quantify, such as emotional intelligence and creativity.  Finally, although the evaluation framework is designed to be comprehensive, the interpretation of outcomes may still be influenced by subjective decisions made during the task selection and formulation process. These limitations should be carefully considered, and future iterations of the benchmark could benefit from addressing these concerns to enhance its applicability and robustness.


\section*{Ethical Considerations}
We discuss the following ethical considerations related to our \texttt{NUMINA} dataset: (1) \textbf{Intellectual Property.} The ScanQA dataset is distributed with the CC BY-NC-SA 3.0 license\footnote{\url{https://creativecommons.org/licenses/by-nc-sa/3.0/}}, and the ScanNet dataset is also available for research use\footnote{\url{https://github.com/ScanNet/ScanNet}}. (2) \textbf{Worker Treatments.} We hired five annotators and fairly pay them according to agreed salaries and workloads. (3) \textbf{Intended Use.} \texttt{NUMINA} can be utilized to develop more persuasive 3D reasoning models. Researchers can also inherit our dataset design and utilize \texttt{NUMINA-Flow} to develop their own datasets. (4) \textbf{Controlling Potential Risks.} Since the documents of \texttt{NUMINA} do not contain private information and annotating this dataset is not necessary to make many judgements about social risks, we believe \texttt{NUMINA} does not introduce any additional risks. We manually verified some randomly sampled data to ensure the dataset did not contain risky issues. (5) \textbf{AI Assistance.} The writing of this paper adopts ChatGPT in refining some sentences.

\bibliography{latex/reference}

\appendix

\section{Templates for Each Category} \label{sec:templates}
Table \ref{tab:templates} presents the QA-pair template for each task type in \texttt{NUMINA}. In this framework, <OBJX> within the question is substituted with actual object names present in the given scenario, while the corresponding answer is generated based on the annotated dataset. This ensures consistency and accuracy in the question-answer pairs while maintaining adaptability across different scenarios.

Each category assesses different aspects of an AI model’s reasoning capabilities related to object properties, spatial relationships, and numerical estimations. Fact Validation (highlighted in red) focuses on binary (yes/no) reasoning about object comparisons, including quantity differences, relative distances, and bounding box size comparisons. Prompt Matching (highlighted in green) evaluates the model’s ability to correctly interpret and match prompts to predefined categorical choices, such as identifying an object's color, spatial positioning, or orientation. Numerical Inference (highlighted in blue) measures the model’s ability to provide precise numerical responses, including object counting, distance estimation, and volume computation. These templates serve as standardized benchmarks to assess LLMs' spatial reasoning, object recognition, and numerical estimation accuracy in indoor environments.

\begin{table*}[t]
    \centering
    \resizebox*{\linewidth}{!}{
    \begin{tabular}{|l|}
    \hline
        \textcolor{red}{\textbf{Fact Validation}} \\
        \textcolor{red}{\textbf{Q1:} Are there fewer <OBJ1> than <OBJ2>? Please reply with a ``yes'' or ``no'' only.} \\
        \textcolor{red}{\textbf{A1:} yes} \\
        \textcolor{red}{\textbf{Q2:} Is the distance between <OBJ1> and <OBJ2> greater than the distance between <OBJ3> and <OBJ4>?} \\
        \textcolor{red}{\textbf{A2:} no} \\
        \textcolor{red}{\textbf{Q3:} Is the size of the bounding box of <OBJ1> less than the one of <OBJ2>? Select ``yes'' or ``no'' as the answer.} \\
        \textcolor{red}{\textbf{A3:} yes} \\
        
    \hline
        \textcolor{green}{\textbf{Prompt Matching}} \\
        \textcolor{green}{\textbf{Q1:} What color is the <OBJ1>? A) Red B) Blue C) Green D) Yellow E) Black} \\
        \textcolor{green}{\textbf{A1:} B} \\
        \textcolor{green}{\textbf{Q2:} What is to the left of the <OBJ1>? A) Armchair B) Bookshelf C) Lamp D) Painting E) Couch} \\
        \textcolor{green}{\textbf{A2:} E} \\
        \textcolor{green}{\textbf{Q3:} What is the  <OBJ1> facing? A) Bookshelf B) Window C) Door D) Round table E) Fireplace} \\
        \textcolor{green}{\textbf{A3:} D} \\
    \hline
    \textcolor{blue}{\textbf{Numerical Inference}} \\
        \textcolor{blue}{\textbf{Q1:} Please count the number of <OBJ1> in the room. Give a number as the answer.} \\
        \textcolor{blue}{\textbf{A1:} 2} \\
        \textcolor{blue}{\textbf{Q2:} Please estimate the distance between the <OBJ1> and <OBJ2> in the room in meters. Give a numerical response.} \\
        \textcolor{blue}{\textbf{A2:} 1.04} \\
        \textcolor{blue}{\textbf{Q3:} Can you estimate the volume of the bounding box of <OBJ1> in cubic meters? Give a numerical response.} \\
        \textcolor{blue}{\textbf{A3:} 3.83} \\
    \hline
    
    \end{tabular}
    }
    \caption{Templates for each category in \texttt{NUMINA}.}
    \label{tab:templates}
\end{table*}

\section{Details of Dataset Construction} \label{sec:details}
The construction of the dataset follows a dual-track generation pipeline, \texttt{NUMINA-Flow}, which comprises independent LLM-based and rule-based methods. 
Each track operates separately, contributing contrapositive strengths to ensure high-quality, factually grounded question-answer pairs. 
The LLM-based track focuses on rewriting Short-Answer Questions (SAQs) from ScanQA into structured formats, while the rule-based track generates questions directly from annotated scene data. Together, these methods enhance linguistic fluency, maintain logical consistency, and minimize hallucinations.

\subsection{LLM-Based Generation}
The LLM-based track transforms SAQs into other question types, including Prompt Matching (PM) and Fact Validation (FV) Questions. 
This process leverages the generative capabilities of large language models while enforcing structural constraints to ensure output consistency and reliability.
Key components of this track include:

\begin{itemize}[itemindent=0em,itemsep=0em]
    \item \textbf{Structured Prompting:} 
    The LLM is guided by carefully designed prompts that explicitly define the rewriting task, specify output constraints in JSON format, and provide illustrative examples. 
    These constraints ensure that the reformulated questions maintain clarity and coherence while adhering to strict formatting guidelines.
    
    \item \textbf{Rigorous Validation:} 
    Post-generation, an automated validation process verifies key structural properties, such as the correct number of options, proper answer placement, and logical distinctiveness of distractors (for PM) or alternative statements (for FV). 
    Any outputs failing these criteria are discarded and regenerated to maintain dataset integrity.
\end{itemize}

To ensure consistency and high-quality question generation, the system employs predefined prompts tailored for different question types. Tables
\ref{tab:llm_pm_prompt} and \ref{tab:llm_fv_prompt} illustrate the structured LLM prompts used for PM and FV generation, respectively.

\begin{table*}[]
    \centering
    \resizebox*{\linewidth}{!}{
    \begin{tabularx}{\linewidth}{|X|}
    \hline
        \textbf{Introduction} \\  
        Please rewrite a Short Answer Question (SAQ) into a Prompt Matching (PM) format with \texttt{n} options. 
    \\\hline
        \textbf{Input} \\  
        - Original SAQ: \textcolor{blue}{<Original Question>} \\  
        - Original Answer: \textcolor{blue}{<Original Answer>} \\  
        - Expected Correct Option: \textcolor{blue}{<Correct Option Label>} \\  
        - Number of Options: \textcolor{blue}{n} 
    \\\hline
        \textbf{Task Description} \\  
        1. Convert the SAQ into a clear and concise PM format. \\  
        2. Generate \textcolor{blue}{n-1} incorrect options as distractors: \\  
        \quad - Keep distractors in the same category as the correct answer (e.g., if the answer is a noun, all distractors should be nouns). \\  
        \quad - Ensure distractors are plausible but incorrect. \\  
        \quad - Avoid synonyms, overly obvious wrong choices, or options that reveal the correct answer. \\  
        3. Place the correct answer exactly as specified (including any spelling variations) in the designated option label. \\  
        4. Format the options as follows: \\  
        \quad - \texttt{``A) Option A  B) Option B  C) Option C ...''} \\  
        \quad - Ensure the correct answer appears at the predefined correct option label. \\  
        5. Include a hint instructing the user to answer using the correct option letter. 
    \\\hline
        \textbf{Output Format} \\  
        Return only a JSON object with the following keys: \\  
        \{ \\  
        \quad \texttt{``question''}: ``The rewritten PM'', \\  
        \quad \texttt{``Answer''}: ``The correct option label (e.g., 'A')'' \\  
        \} \\  
        Do not include any additional text or explanations outside this JSON object.  
    \\\hline
        \textbf{Example Formatting} \\  
        \textbf{Input:} \\
        - SAQ: {``What is the capital of France?''} \\  
        - Answer: {``Parris''} (note the misspelling) \\  
        - Expected Correct Option: {``B''} \\  
        - Number of Options: {4} \\  

        \textbf{Expected Output:} \\  
        \{ \\  
        \quad \texttt{``question''}: ``What is the capital of France? Answer using the correct option letter. \\  
        \quad A) Berlin  B) Parris  C) London  D) Rome'', \\  
        \quad \texttt{``Answer''}: ``B'' \\  
        \} 
    \\\hline
    \end{tabularx}
    }
    \caption{LLM Prompt for PM question Generation.}
    \label{tab:llm_pm_prompt}
\end{table*}

\begin{table*}[]
    \centering
    \resizebox*{\linewidth}{!}{
    \begin{tabularx}{\linewidth}{|X|}
    \hline
        \textbf{Introduction} \\  
        Please rewrite a Short Answer Question (SAQ) into a Fact Validation (FV) format and also generate its contrapositive version.  
    \\\hline
        \textbf{Input} \\  
        - Original SAQ: \textcolor{blue}{<Original Question>} \\  
        - Original Answer: \textcolor{blue}{<Original Answer>} \\  
        - Boolean Indicator: \textcolor{blue}{<True or False>} \\  
        - Answer Options: \textcolor{blue}{``<Affirmative Word>''} (for True) / \textcolor{blue}{``<Negative Word>''} (for False)  
    \\\hline
        \textbf{Task Description} \\  
        1. Convert the SAQ into a clear, factual statement incorporating the given answer. \\  
        2. Rewrite the statement into a FV based on the Boolean Indicator: \\  
        \quad - If \texttt{True} → Keep the statement affirmative (e.g., \texttt{``Bob sits next to Alice.''}). \\  
        \quad - If \texttt{False} → Negate the statement (e.g., \texttt{``Bob does not sit next to Alice.''}). \\  
        3. Append the answer options at the end as: \\  
        \quad - ``Is this correct? Answer with \textcolor{blue}{<Affirmative Word>} or \textcolor{blue}{<Negative Word>}.'' \\  
        4. Generate a contrapositive question by logically inverting the original statement and answer. \\  
        5. Ensure that both original and contrapositive questions are logically and grammatically correct.  
    \\\hline
        \textbf{Output Format} \\  
        Return only a JSON object with the following keys: \\  
        \{ \\  
        \quad \texttt{``question''}: ``The rewritten FV question'', \\  
        \quad \texttt{``Answer''}: ``The answer option corresponding to the preset Boolean indicator'', \\  
        \quad \texttt{``cp\_question''}: ``The contrapositive version of the FV'', \\  
        \quad \texttt{``cp\_answer''}: ``The answer option corresponding to the negation of the preset Boolean indicator'' \\  
        \} \\  
        Do not include any additional text or explanations outside this JSON object.  
    \\\hline
        \textbf{Example Formatting} \\  
        \textbf{Input:} \\  
        - SAQ: {``Who sits next to Alice?''} \\  
        - Answer: {``Bob''} \\  
        - Boolean Indicator: {False} \\  
        - Answer Options: {``\textcolor{blue}{<Affirmative Word>}''} (True) / {``\textcolor{blue}{<Negative Word>}''} (False) \\  

        \textbf{Expected Output:} \\  
        \{ \\  
        \quad \texttt{``question''}: ``Bob does not sit next to Alice. Is this correct? Answer with \textcolor{blue}{<Affirmative Word>} or \textcolor{blue}{<Negative Word>}.'', \\  
        \quad \texttt{``Answer''}: ``\textcolor{blue}{<Negative Word>}'', \\  
        \quad \texttt{``cp\_question''}: ``Bob sits next to Alice. Is this correct? Answer with \textcolor{blue}{<Affirmative Word>} or \textcolor{blue}{<Negative Word>}.'', \\  
        \quad \texttt{``cp\_answer''}: ``\textcolor{blue}{<Affirmative Word>}'' \\  
        \}  
    \\\hline
    \end{tabularx}
    }
    \caption{LLM Prompt for FV Generation.}
    \label{tab:llm_fv_prompt}
\end{table*}

\subsection{Rule-Based Generation}

Independent of the LLM-based track, the rule-based track constructs questions directly from structured scene annotations using deterministic procedures. 
This method ensures factual accuracy by deriving question-answer pairs solely from observable, verifiable data.
The rule-based track consists of the following steps:

\begin{itemize}[itemindent=0em,itemsep=0em]
    \item \textbf{Candidate Selection:} 
    The pipeline selects objects or object pairs that meet predefined criteria, such as bounding box sizes, or numerical attributes. 
    Non-informative labels (e.g., \texttt{item}, \texttt{object}) are filtered out to ensure relevance and diversity in the generated questions.

    \item \textbf{Automated Question Formation:} 
    Once appropriate candidates are identified, the pipeline applies structured templates and logical rules to generate FV or NI questions. 
    These templates guarantee consistency and accuracy in queries while keeping them diverse and reflective of real-world data.
    
    \item \textbf{Grounding in Factual Data:} 
    Since this track draws directly on scene metadata (e.g., object labels, spatial relationships, and numerical attributes), the resulting questions are inherently grounded in real, observable data, minimizing ambiguity and eliminating the risk of hallucinated content.
    
\end{itemize}

By operating as independent yet contrapositive tracks, the LLM-based and rule-based approaches provide a balanced mechanism for dataset construction. 
The LLM-based track enhances linguistic naturalness while adhering to strict structural constraints, whereas the rule-based track guarantees factual accuracy through deterministic question generation. 
This dual-track framework ensures that the resulting dataset is both diverse and reliable, making it well-suited for evaluating AI models in spatial reasoning and numerical inference tasks.

\section{System Prompt} \label{sec:system prompt}
Table \ref{tab:system_prompt} presents the system prompt provided to the LLM, which defines its role, outlines the task requirements, and specifies the meanings of ``distance'' and ``volume'' used in the subsequent questions. This ensures standardization in spatial reasoning tasks within the \texttt{NUMINA} dataset, promoting uniform interpretation and evaluation of distance and volume-related queries. However, the LLM struggles to accurately comprehend the concept of convex hull distance, likely due to the complexity of its calculation. As a result, the model performs poorly on all distance-related tasks, highlighting its limitations in numerical spatial reasoning.

\begin{table*}[]
    \centering
    \resizebox*{\linewidth}{!}{
    \begin{tabularx}{\linewidth}{|X|}
    \hline
        You are an AI Assistant providing accurate, detailed, and polite explanations. Your 
        goal is to give clear and helpful responses. Distances refer to convex hull distances,  
        the shortest distance between points on the convex hulls of two objects, where a convex 
        hull is the smallest convex shape enclosing an object. Volumes refer to the space within 
        an object's minimum axis-aligned bounding box (AABB), a rectangular box aligned 
        with the coordinate axes. Always use these definitions for distances and volumes. 
    \\\hline
    \end{tabularx}
    }
    \caption{System Prompt of \texttt{NUMINA}.}
    \label{tab:system_prompt}
\end{table*}

\section{Logical Consistency Question-Answer Pairs} \label{sec:lc}
\begin{table*}[]
    \centering
    \resizebox*{\linewidth}{!}{
    \begin{tabularx}{\linewidth}{|X|}
    \hline
       
        \textbf{Ori-Q:} Are there \textcolor{blue}{fewer} kitchen counter than table? Reply with a ``yes'' or ``no'' only. \\
        \textbf{Ori-A:} \textcolor{red}{yes} \\
        \textbf{CP-Q:} Can you tell if the count of kitchen counter is \textcolor{blue}{greater than or equal} to the count of table? Reply with a ``yes'' or ``no'' only. \\
        \textbf{CP-A:} \textcolor{red}{no} \\
        
    \hline
        
        \textbf{Ori-Q:} Is the distance between tissue box and guitar \textcolor{blue}{greater than} the one between refrigerator and sink? Offer a ``yes'' or ``no'' as the answer. \\
        \textbf{Ori-A:} \textcolor{red}{no} \\
        \textbf{CP-Q:} Is the distance between tissue box and guitar \textcolor{blue}{less than or equal} to the distance between refrigerator and sink? Offer a ``yes'' or ``no'' as the answer. \\
        \textbf{CP-A:} \textcolor{red}{yes} \\
        
    \hline
    
        \textbf{Ori-Q:} Is the size of the bounding box of kitchen counter \textcolor{blue}{greater than or equal} to the one of door? Select ``yes'' or ``no'' as the answer. \\
        \textbf{Ori-A:} \textcolor{red}{yes} \\
        \textbf{CP-Q:} Can you tell if the volume of the bounding box of kitchen counter is \textcolor{blue}{less than} the volume of the bounding box of door? Select ``yes'' or ``no'' as the answer. \\
        \textbf{CP-A:} \textcolor{red}{no} \\
        
    \hline
    
    \end{tabularx}
    }
    \caption{Templates for logical consistency QA-Pairs in \texttt{NUMINA}. Each original question (Ori-Q) is followed by its corresponding original answer (Ori-A), while a logically equivalent or contrapositive question (CP-Q) is provided to verify coherence in reasoning, with its expected logically consistent answer (CP-A). such contrary-logical QA-pairs aids in evaluating the reliability and robustness of language models in maintaining logical consistency.}
    \label{tab:cp-templates}
\end{table*}

Table \ref{tab:cp-templates} presents a structured set of logical consistency QA-pairs in \texttt{NUMINA}, designed to assess the reasoning capabilities of language models. Each original question (Ori-Q) is accompanied by its corresponding original answer (Ori-A), establishing a baseline response. Additionally, a contrapositive or logically equivalent question (CP-Q) is introduced to verify coherence in reasoning, with its expected logically consistent answer (CP-A).

The QA pairs encompass comparisons involving quantities, distances, and bounding box sizes, where the contrapositive questions are framed to test whether the model maintains consistency when restating or reversing relational statements (e.g., ``\textit{greater than}'' vs. ``\textit{less than or equal to}''). This framework ensures the reliability and robustness of language models by systematically evaluating their logical consistency across related queries.

\section{Chain of Thoughts Question-Answer Pairs} \label{sec:cot}
Table \ref{tab:cot-templates} presents a set of question-answer (QA) pairs designed to illustrate the application of Chain-of-Thought (CoT) reasoning in the \texttt{NUMINA} dataset. Each original question (Ori-Q) is accompanied by a direct answer (Ori-A), followed by a reformulated question incorporating CoT reasoning (CoT-Q) and its corresponding answer (CoT-A). The CoT-based responses provide step-by-step logical explanations to enhance interpretability and reasoning transparency.

The QA pairs cover various spatial reasoning and numerical estimation tasks, including volume computation, object counting, and comparative distance assessments. For instance, in the volume estimation task, the CoT response explicitly details the bounding box dimensions and the multiplication process to derive the final volume, reinforcing numerical reasoning. Similarly, in object counting tasks, the CoT response explicitly states the numerical values being compared before concluding the answer. Distance-based reasoning follows a similar pattern, where spatial distances are quantitatively analyzed before providing the final decision.

\begin{table*}[]
    \centering
    \resizebox*{\linewidth}{!}{
    \begin{tabularx}{\linewidth}{|X|}
    \hline
       
        \textbf{Ori-Q:} Can you estimate the volume of the bounding box of bed in cubic meters? Give a numerical response. \\
        \textbf{Ori-A:} \textcolor{red}{3.83} \\
        \textbf{CoT-Q:} Can you estimate the volume of the bounding box of bed in cubic meters? Please solve the problem step by step.
        Show each intermediate thought process clearly and provide the final answer after completing the reasoning process. \\
        \textbf{CoT-A:} Given the bounding box dimensions of the object along the X, Y, and Z axes as 1.98 m, 2.32 m, and 0.83 m respectively, the volume of the bounding box is calculated as (length x width x height) yielding approximately \textcolor{red}{3.83} cubic meters. \\
        
    \hline
        
        \textbf{Ori-Q:} Are there fewer stool than cabinet? Please reply with a ``yes'' or ``no'' only. \\
        \textbf{Ori-A:} \textcolor{red}{yes} \\
        \textbf{CoT-Q:} Are there fewer stool than cabinet? Please solve the problem step by step. Show each intermediate thought
        process clearly and provide the final answer after completing the reasoning process. \\
        \textbf{CoT-A:} Given the count of stool as 4 and the count of cabinet as 5, the count of stool is less than the count of cabinet. Therefore, the answer is \textcolor{red}{yes}. \\
        
    \hline
    
        \textbf{Ori-Q:} Is backpack and toaster further than coffee table and mirror? Please provide a ``yes'' or ``no'' as the answer. \\
        \textbf{Ori-A:} \textcolor{red}{no} \\
        \textbf{CoT-Q:} Is backpack and toaster further than coffee table and mirror? Please solve the problem step by step. Show each intermediate thought process clearly and provide the final answer after completing the reasoning process. \\
        \textbf{CoT-A:} The distance between backpack and toaster is approximately 5.05 meters. The distance between coffee table and mirror is approximately 5.27 meters. Since the distance between backpack and toaster is greater than the distance between coffee table and mirror, the correct answer is \textcolor{red}{no}. \\
        
    \hline

        \textbf{Ori-Q:} Is the size of the bounding box of curtain approximately equal to the one of cabinet? Kindly provide a ``yes'' or ``no'' as the answer. \\
        \textbf{Ori-A:} \textcolor{red}{no} \\
        \textbf{CoT-Q:} Is the size of the bounding box of curtain approximately equal to the one of cabinet? Please solve the problem step by step. Show each intermediate thought process clearly and provide the final answer after completing the reasoning process. \\
        \textbf{CoT-A:} Given the volume of the bounding box of curtain as 10.41 cubic meters and the volume of the bounding box of cabinet as 4.88 cubic meters, the volume of the bounding box of curtain is approximately equal to the volume of the bounding box of cabinet. Therefore, the answer is \textcolor{red}{no}. \\
        
    \hline
    
    \end{tabularx}
    }
    \caption{Templates for Chain-of-Thoughts QA-Pairs in \texttt{NUMINA}. Each example includes an original question (Ori-Q) with its corresponding direct answer (Ori-A), followed by a CoT-augmented question (CoT-Q) that prompts a step-by-step reasoning process. The CoT-augmented answer (CoT-A) then provides a detailed breakdown of the logical steps leading to the final response.}
    \label{tab:cot-templates}
\end{table*}

\end{document}